\documentclass[letterpaper, 10 pt, conference]{ieeeconf}  
\pdfoutput=1

\IEEEoverridecommandlockouts                              

\usepackage{amsmath} 
\usepackage{gensymb}
\usepackage{textcomp}
\usepackage{mathtools}
\usepackage{xargs}  
\usepackage{units}
\usepackage{pgfplots}
\usepackage{caption}
\usepackage{subcaption}
\usepackage{graphicx,xcolor}
\usepackage{colortbl}
\usepackage[T1]{fontenc}

\usepackage{amssymb}  
\usepackage{xfrac}
\usepackage{mathtools}
\usepackage{mathrsfs}
\usepackage{tikzsymbols}
\usepackage[binary-units=true]{siunitx}
\usepackage{bm}

\usepackage[bordercolor=white,backgroundcolor=gray!30,linecolor=black,colorinlistoftodos]{todonotes}
\newcommandx{\tj}[1]{\todo[color=red, inline]{tj: {#1}}}

\makeatletter
\let\NAT@parse\undefined
\makeatother
\usepackage{hyperref}

\title{\LARGE \bf
Nonlinear Model Predictive Guidance for Fixed-wing UAVs Using Identified Control Augmented Dynamics
}

\author{Thomas Stastny and Roland Siegwart
\thanks{All authors are with the Swiss Federal Institute of Technology (ETH) Z\"urich, Autonomous Systems Lab (www.asl.ethz.ch), Leonhardstrasse 21, 8092 Z\"urich, Switzerland.
{\tt \{\underline{firstname.lastname}}\}@mavt.ethz.ch.}
}
\usepackage{xcolor}

\newcommand{\subparagraph}{}
\usepackage{titlesec}

\begin{document}

\maketitle
\thispagestyle{plain}
\pagestyle{plain}

\setlength{\floatsep}{4pt plus 2pt minus 2pt}
\setlength{\textfloatsep}{4pt plus 2pt minus 2pt}

\begin{abstract}
As off-the-shelf (OTS) autopilots become more widely available and user-friendly and the drone market expands, safer, more efficient, and more complex motion planning and control will become necessary for fixed-wing aerial robotic platforms.
Considering typical low-level attitude stabilization available on OTS flight controllers, this paper first develops an approach for modeling and identification of the control augmented dynamics for a small fixed-wing Unmanned Aerial Vehicle (UAV).
A high-level Nonlinear Model Predictive Controller (NMPC) is subsequently formulated for simultaneous airspeed stabilization, path following, and soft constraint handling, using the identified model for horizon propagation.
The approach is explored in several exemplary flight experiments including path following of helix and connected Dubins Aircraft segments in high winds as well as a motor failure scenario.
The cost function, insights on its weighting, and additional soft constraints used throughout the experimentation are discussed.
\end{abstract}


\section{INTRODUCTION}

As off-the-shelf (OTS) autopilots become more widely available and user-friendly and the drone market expands, safer, more efficient, and more complex motion planning and control will become necessary for aerial robotic platforms.
Tools for auto-code-generation of fast, efficient embedded nonlinear solvers, e.g. ACADO Toolkit~\cite{Houska2011a} or FORCES\footnote{\url{https://www.embotech.com/FORCES-Pro}}, are becoming popular for the high-level control design of such systems.
Exemplary applications of these tools, using Nonlinear Model Predictive Control (NMPC), have been experimentally shown on multi-copters for various high-level tasks such as trajectory tracking~\cite{Kamel2017_ROSBOOK}, inter-vehicle collision avoidance~\cite{Kamel2017_RobustCollisionAvoidMAVNMPC}, and aerial manipulation~\cite{Lunni2017_NMPCAerialManipulation}.
The NMPC formulation conveniently offers the capability to solve receding horizon optimal control problems with consideration of nonlinear dynamics and handling of state/input constraints, a valuable set of functionalities for flying platforms aiming to satisfy the ever-increasing complexity of desired autonomous behaviors.
For large-scale sensing and mapping applications, small fixed-wing unmanned aerial vehicles (UAVs) provide advantages of longer range and higher speeds than rotorcraft.
However, unlike their multi-copter counterparts, experimental implementation and validation of NMPC approaches on fixed-wing platforms is almost non-existent.
To examine the state-of-the-art in fixed-wing specific NMPC formulation, one must consider simulation studies within the literature.
High-level guidance formulations, using two-degrees-of-freedom (2DOF) kinematic models for horizon propagation, have been shown for the 2D path following case~\cite{kang2006design,kang2009linear,yang2009adaptive}, and with 3DOF kinematic models for 3D soaring~\cite{Liu_NMPC_AutoSoaring} or automatic landing~\cite{Joos2011_NMPCSimAutoLand}. 
Other works have considered lower-level formulations, either incorporating all objectives from obstacle avoidance to actuator penalty directly~\cite{Gros2012_AircraftFastNMPC}, focusing on low-level states only, e.g. for deep-stall landing~\cite{Mathisen2016_NMPCPrecisionDeepStallLand}, or augmenting the internal low-level model with guidance logic~\cite{Stastny_JDSMC2014,Stastny_GNC2015}.
\begin{figure}[t!]
\centering
\includegraphics[width=\linewidth]{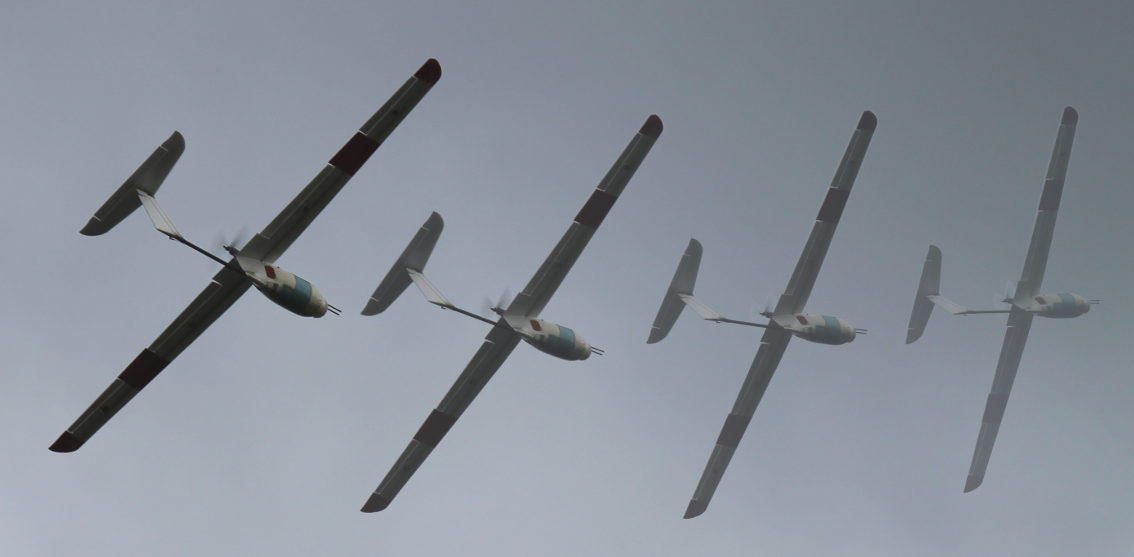}
\caption{Techpod, fixed-wing unmanned aerial test platform.}
\label{fig:techpod}
\end{figure}
Higher-level formulations typically utilize simple parameterless kinematic models, assuming that lower-level controllers adequately track high-level commands.
These approaches rarely consider details of integration with increasingly ubiquitous OTS autopilots and their low-level control structures.
On the other hand, lower-level formulations, if implemented on real aircraft, require extensive wind tunnel testing and/or flight experimentation for actuator-level aerodynamic system identification (ID), a time consuming and potentially safety-critical process.
In our previous work~\cite{Stastny_GNC2017}, we first explored the concept of encapsulating the closed-loop autopilot roll channel response dynamic into the internal model of a high-level lateral-directional NMPC, taking a `middle road' between full classical ID and model-free formulations.
Broader application of this approach to 3D problems requires extending the control augmented modeling to a full, coupled lateral-directional and longitudinal structure; an extension we provide in the present work.
In this paper, we will first develop an approach to modeling and identification of \emph{control augmented dynamics} for a conventional fixed-wing platform with a widely available OTS autopilot in the loop, utilizing a standard sensor suite.
We will secondly detail a high-level NMPC cost function design for simultaneous airspeed stabilization, path following, and soft constraint handling, utilizing the identified model internally.
We take special consideration of practical implementation insights throughout this work, such as explicit consideration of high winds as well as on-board computational constraints, and conclude with a set of representative flight experiments for validation of the approach.
%


\section{CONTROL AUGMENTED MODELING}
\begin{figure}
	\centering
	\includegraphics[width=\linewidth,trim={0.6cm 0.5cm 0cm 0cm},clip]{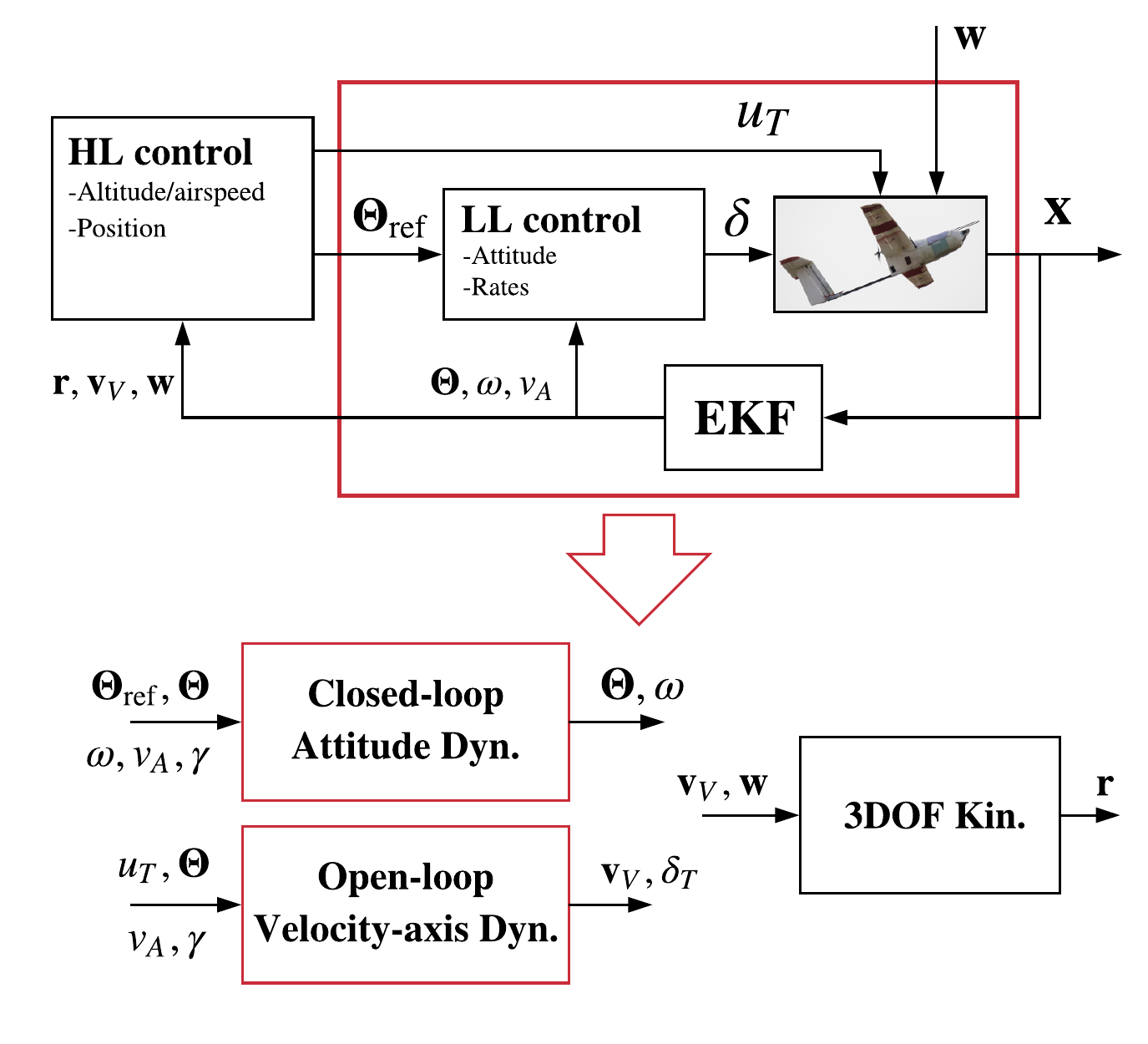}
	\caption{Model abstraction of the closed-loop attitude dynamics, open-loop velocity-axis dynamics, and 3DOF kinematics.}
	\label{fig:cascaded_modeling}
\end{figure}

\begin{figure}
	\centering
	\includegraphics[trim={2.5cm 2.5cm 0cm 1.5cm},clip,width=\linewidth]{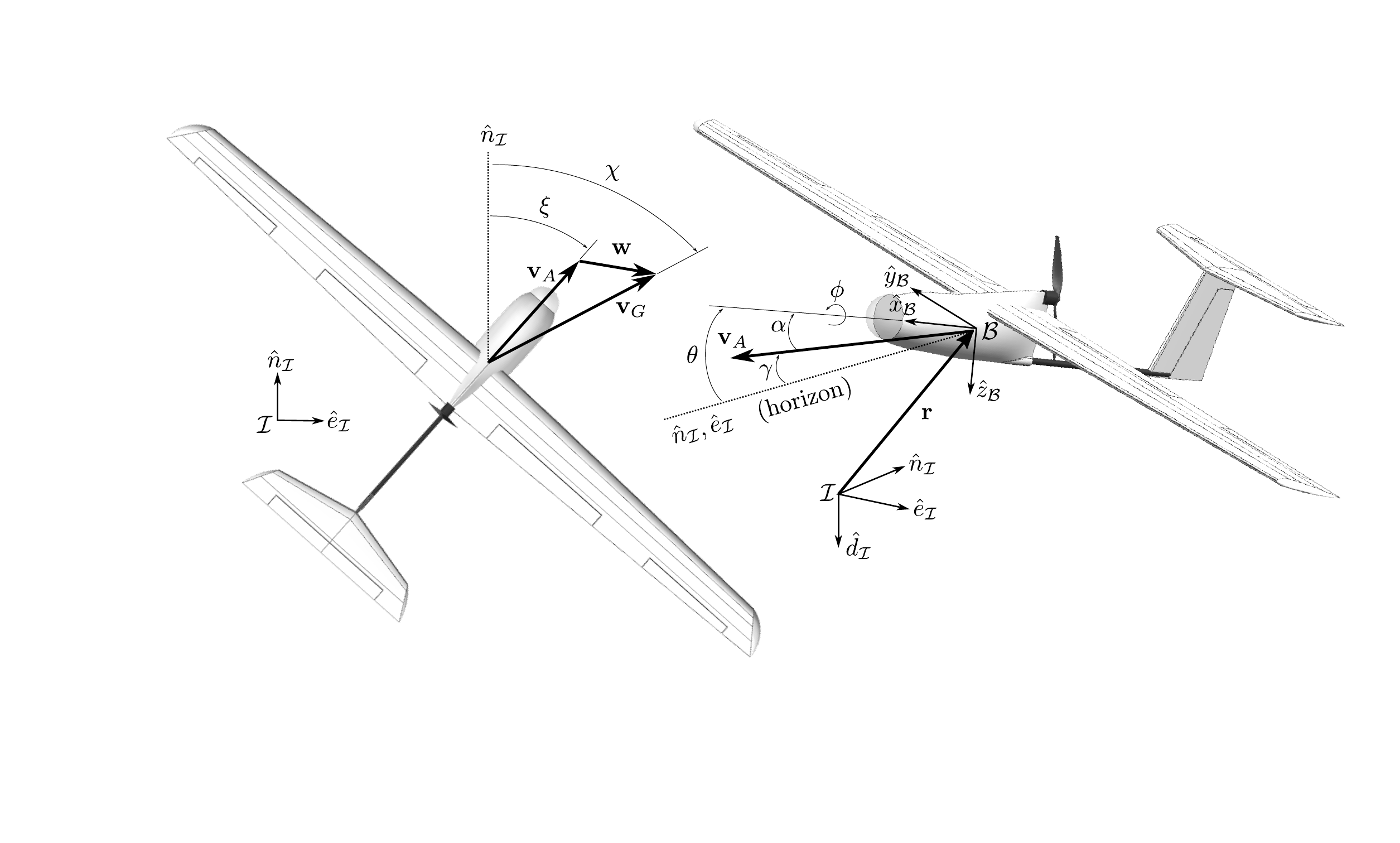}
	\caption{Inertial $\mathcal{I}$ and body $\mathcal{B}$ axes and state definitions.}
	\label{fig:axes_definitions}
\end{figure}

%
A typical fixed-wing system/control architecture is shown in Fig.~\ref{fig:cascaded_modeling}, e.g. similarly implemented in open-source autopilot firmwares \emph{PX4}\footnote{\url{http://pixhawk.org}} and \emph{Ardupilot}\footnote{\url{ardupilot.org}}.
A low-level (LL) control structure runs on the pixhawk microcontroller consisting of a cascaded PD attitude control/rate damping approach with coordinated turn feed-forward terms, using both rudder and elevator compensation, as well as dynamic pressure scaling on control actuation, see~\cite{Oettershagen2016_JFR} for more details.
High-level (HL) lateral guidance, e.g. within \emph{PX4} firmware $\mathcal{L}_1$-guidance~\cite{park2007performance}, steers the aircraft position and velocity toward waypoints or paths by commanding roll angle references $\phi_\text{ref}$, and airspeed/altitude are controlled, e.g. via Total Energy Control System (TECS) \cite{tecs}, by commanding pitch angle references $\theta_\text{ref}$ and throttle input $u_T$.
To replace this high-level module with a unified model predictive controller, one must characterize the underlying dynamics.
Here, we propose a ``cascaded" modeling approach, defining two low-level model structures as grey-box models: 1) the stabilized, closed-loop attitude dynamics, eq.~\eqref{eq:CL_dynamics}, and 2) the open-loop velocity-axis dynamics, eq.~\eqref{eq:OL_dynamics}. Their outputs are fed to the standard (parameterless) 3DOF kinematic equations~\eqref{eq:3DOF_kinematics}.
State/axes definitions may be seen in Fig.~\ref{fig:axes_definitions}.
\subsection{Closed-loop attitude dynamics}
We model the input-output relationship of the closed-loop LL attitude controlled system in eq.~\eqref{eq:CL_dynamics}; specifically, how the attitude and body rates respond to attitude references.
%
%
%
The structure contains coupled lateral-directional and longitudinal states and parameters as well as nonlinearities, particularly owing to the longitudinal effects.
LL controllers are often tuned for one, or very few, trim conditions around the standard flight operation point, allowing the control augmented behavior throughout the flight envelope to vary, further motivating included nonlinear airspeed dependence.
\begin{equation}
\label{eq:CL_dynamics}
\left(\begin{matrix}
\dot{\phi}\\
\dot{\theta}\\
\dot{p}\\
\dot{q}\\
\dot{r}
\end{matrix}\right)=\left(\begin{matrix}
p\\
q\cos\phi-r\sin\phi\\
l_p p + l_r r + l_{e_\phi} \left(\phi_\text{ref} - \phi\right)\\
{v_A}^2\left(m_0 + m_\alpha \alpha + m_q q + m_{e_\theta} \left(\theta_\text{ref} - \theta\right)\right)\\
n_r r + n_\phi \phi + n_{\phi_\text{ref}} \phi_\text{ref}
\end{matrix}\right)
\end{equation}
\noindent where $\mathbf{\Theta}=\left[\phi,\theta\right]$ is the aircraft attitude (roll and pitch, respectively), $\bm{\omega}=\left[p,q,r\right]^T$ are the body roll, pitch, and yaw rates, respectively, $v_A$ is the airspeed (air-mass relative), and $\bm{\varphi}_\text{CL}=\left[l_p,l_r,l_{e_\phi},m_0,m_\alpha,m_q,m_{e_\theta},n_r,n_\phi,n_{\phi_\text{ref}}\right]^T$ is the set of parameters to identify.
\subsection{Open-loop dynamics}
As airspeed is controlled on a high-level basis within the given autopilot structure, there exists a non-stabilized (open-loop) dynamic from throttle input to UAV outputs we must model, see eq.~\eqref{eq:OL_dynamics}.
\begin{equation}
\label{eq:OL_dynamics}
\left(\begin{matrix}
\dot{v_A}\\
\dot{\gamma}\\
\dot{\xi}\\
\dot{\delta_T}
\end{matrix}\right)=\left(\begin{matrix}
\frac{1}{m}\left(T\cos\alpha-D\right)-g\sin\gamma \\
\frac{1}{m v_A}\left[\left(T\sin\alpha+L\right)\cos\phi - mg\cos\gamma\right] \\
\frac{\sin\phi}{m v_A\cos\gamma}\left(T\sin\alpha+L\right)\\
\left(u_T-\delta_T\right)/\tau_T
\end{matrix}\right)
\end{equation}
\noindent where $m$ is the mass, $g$ is the acceleration of gravity, $\gamma$ is the air-mass relative flight path angle, $\alpha$ is the aircraft angle of attack (AoA), $\delta_T$ is a virtual throttle \emph{state} lagged from input $u_T\in[0,1]$ by time constant $\tau_T$, and $\xi$ is the aircraft heading, defined from North to the airspeed vector.
This 3DOF model is often used as a simplified dynamic formulation in aerospace controls literature, containing only forces, the assumption being that moments are controlled on a lower-level and the overall behavior of the high-level states may be described in a quasi-steady manner.
Further, the model only considers forces in the longitudinal axis, making the assumption that no aerodynamic or thrusting side force is generated, in part due to an assumption that the low-level controller appropriately regulates sideslip.
By neglecting sideslip, which is challenging to observe without a vector airdata probe or alpha-beta vane, we are also able to make the approximate relationship $\alpha\approx\theta-\gamma$ (similarly necessary as $\alpha$ is not directly measured) and that heading angle $\xi$ is assumed equivalent to aircraft yaw angle.
We, however, retain the velocity-axis convention, $\mathbf{v}_V=\left[v_A,\gamma,\xi\right]^T$, for aircraft heading $\xi$ as a means to distinguish between lower- and higher-level modeling descriptions.
The force equations are shown in~\eqref{eq:forces}.
\begin{equation}
\label{eq:forces}
\begin{matrix*}[l]
T=\left(c_{T_1}\delta_T+c_{T_2}{\delta_T}^2+c_{T_3}{\delta_T}^3\right)/v_{\infty_\text{prop}}\\
D=\bar{q}S\left(c_{D_0}+c_{D_\alpha}\alpha+c_{D_{\alpha^2}}\alpha^2\right)\\
L=\bar{q}S\left(c_{L_0}+c_{L_\alpha}\alpha+c_{L_{\alpha^2}}\alpha^2\right)
\end{matrix*}
\end{equation}
\noindent where motor thrust $T$ is modeled as power $\mathscr{P}$, a function of throttle input, over the effective propeller free stream, approximated as $v_{\infty_{prop}}\approx v_A\cos\alpha$.
Lift $L$ and drag $D$ forces are scaled with dynamic pressure $\bar{q}$ and wing surface area $S$.
The elaborated model structure contains grey parameters $\bm{\varphi}_\text{OL}=\left[c_{T_1},c_{T_2},c_{T_3},\tau_T,c_{D_0},c_{D_\alpha},c_{D_{\alpha^2}},c_{L_0},c_{L_\alpha},c_{L_{\alpha^2}}\right]^T$ to be identified.
%

%
\subsection{3DOF kinematics}
Finally, parameterless 3DOF kinematics propagate the position through time in wind:
\begin{equation}
\label{eq:3DOF_kinematics}
\left(\begin{matrix}
\dot{n}\\
\dot{e}\\
\dot{d}\\
\end{matrix}\right)=\left(\begin{matrix}
v_A\cos\gamma\cos\xi + w_n \\
v_A\cos\gamma\sin\xi+ w_e \\
-v_A\sin\gamma + w_d \\
\end{matrix}\right)
\end{equation}
\noindent where $\mathbf{r}=\left[n,e,d\right]^T$ are the inertial frame Northing, Easting, and Down position components, respectively, and $\mathbf{w}=\left[w_n,w_e,w_d\right]^T$ are the inertial frame wind components, modeled as static disturbances.

\section{SYSTEM IDENTIFICATION}\label{sec:sys_id}
\subsection{System overview}
All development and experimentation within this work is conducted on the \SI{2.6}{\meter} wingspan, \SI{2.65}{\kilo\gram}, hand-launchable fixed-wing UAV -- Techpod, see Fig.~\ref{fig:techpod}.
The platform is a standard T-tail configuration, fixed-pitch, pusher propeller integrated with a 10-axis ADIS16448 Inertial Measurement Unit (IMU), u-Blox LEA-6H GPS receiver, and Sensirion SDP600 flow-based differential pressure sensor coupled with a one-dimensional pitot-static tube configuration.
Sensor measurements are fused in a light-weight, robust Extended Kalman Filter (EKF)~\cite{Leutenegger2014_SE} running on board a \emph{Pixhawk} Autopilot (168 MHz Cortex-M4F microcontroller with 192 kB RAM) generating state estimates including a local three-dimensional wind vector, modeled statically with slow dynamics.
\subsection{Data collection and organization}
Data from five approx. \SI{40}{\minute} flight tests was collected containing 72 experiment sets (with 1 or 2 identification maneuvers each) spanning a range of 28 \emph{static}, 35 \emph{dynamic}, and 9 \emph{free-form} preprogrammed maneuvers, covering the operational flight envelope (i.e. $v_A\in\left[11,18\right]$\SI{}{\meter\per\second}, $\phi\in\left[-30,30\right]$\SI{}{\degree}, and $\theta\in\left[-15,15\right]$\SI{}{\degree}), all with active attitude stabilization.
%
A 70-30 percent ratio was used for training and validation groups on the static and dynamic sets (together, \SI{87.5}{\percent} of the total number of sets), while the free-form sets were all held back for a ``testing" group (the remaining \SI{12.5}{\percent}).
\begin{itemize}
\item \emph{Static} experiment sets refer to fixed airspeed $v_A$, throttle input $u_T$, and flight path angle $\gamma$ with no dynamic maneuvering (i.e. constant $\phi_\text{ref}$ and $\theta_\text{ref}$).
\item \emph{Dynamic} experiment sets were conducted at various flight speeds and flight path angles utilizing 2-1-1 step inputs (see~\cite{morelli2003low}) for all $u_T$, $\phi_\text{ref}$, and $\theta_\text{ref}$ to excite the low-level autopilot response dynamics.
\item \emph{Free-form} experiment sets refer to manually commanded attitude references and throttle inputs to the stabilized system in an arbitrary fashion.
\end{itemize}
Note that inputs were applied in both independent and coupled combinations (see Figures~\ref{fig:idValCoupled-2} and~\ref{fig:idValCoupled-1} for an example of a coupled identification maneuver). 
All static and dynamic maneuvers were initialized at trim reference commands for a settling period before enacting the steps.
For repeatable experiments, commands were generated in a mostly automated fashion on-board the pixhawk.
%
A data logging rate of \SI{40}{\hertz} was found sufficient to observe the stabilized dynamic responses within the maneuvers.
Care was taken to fly on windless days, and on-board estimates from the EKF are used within the parameter estimation process without any post processing.

\subsection{Time-domain nonlinear grey-box identification}
The MATLAB System Identification Toolbox (\emph{ver. R2016b}) was used for nonlinear grey-box estimation.
The closed-loop $\bm{\varphi}_\text{CL}$ and open-loop $\bm{\varphi}_\text{OL}$ model parameters were identified in a decoupled manner, focusing the parameters to their respective dynamics and outputs in an attempt to avoid any erroneous cost minimization in the optimization across model structures.
Further, decoupling the identifications allows any future change in low-level attitude control parameters only to require adapting the closed-loop attitude response model, while the quasi-steady open-loop model should not change with respect to slightly varying attitude stabilization.
%
%

\begin{figure}
	\centering
	\includegraphics[width=0.95\linewidth]{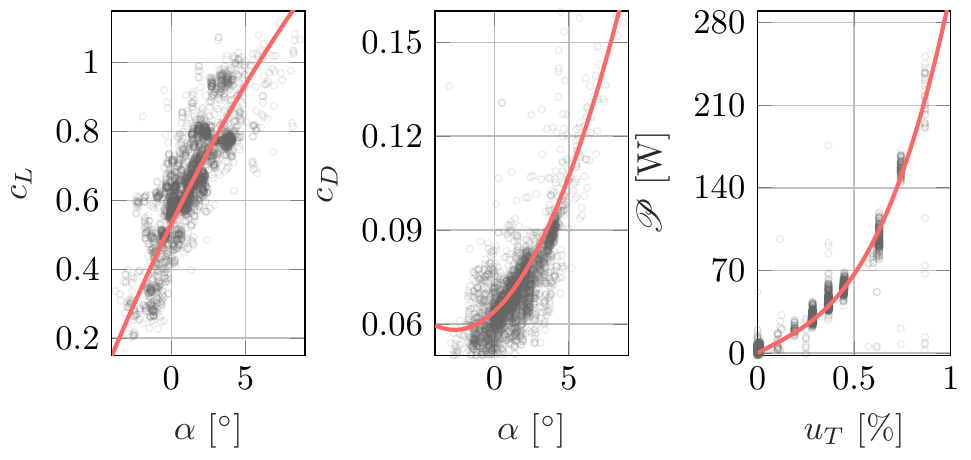}
	\caption{Static aerodynamic and power curves: lift coefficient (left), drag coefficient (center), power (right).
	%
	Acceleration data with corresponding body rates below \SI{1}{\degree\per\second} are displayed. Note this ``sanity check" is important during the identification and model selection process, as the output-error method can easily misrepresent the underlying physics, despite obtaining a low-cost fit.}
	\label{fig:static_curves}
\end{figure}

%
The grey-box structure for the closed-loop attitude dynamics contains states $\mathbf{x}_\text{CL}=\left[\mathbf{\Theta}^T,\bm{\omega}^T\right]^T$, inputs $\mathbf{u}_\text{CL}=\left[\mathbf{\Theta}_\text{ref}^T,v_A,\gamma\right]^T$, and outputs for error minimization $\mathbf{y}_\text{CL}=\left[\mathbf{\Theta}^T,\bm{\omega}^T\right]^T$, and dynamic equations~\eqref{eq:CL_dynamics}.
Note the airspeed and flight path angles are input from the logged data, and not propagated within the model structure.
The grey-box structure for the open-loop dynamics contains states $\mathbf{x}_\text{OL}=\left[v_A,\gamma,\delta_T\right]^T$, dynamic equations~\eqref{eq:OL_dynamics}, inputs $\mathbf{u}_\text{OL}=\left[\mathbf{\Theta}^T,u_T\right]^T$, and outputs $\mathbf{y}_\text{OL}=\left[v_A,\gamma,a_x,a_z\right]^T$, where $a_x$ and $a_z$ are the $x$-body and $z$-body axis accelerations, related to the internal model states as:
\begin{equation}\label{eq:accel}
\left(\begin{matrix}
a_x\\
a_z\end{matrix}\right)=\left(\begin{matrix}
\cos\alpha & \sin\alpha \\
\sin\alpha & -cos\alpha
\end{matrix}\right)\left(\begin{matrix}
\left(T\cos\alpha-D\right)/m \\
\left(T\sin\alpha+L\right)/m
\end{matrix}\right)
\end{equation}
The minimization of body acceleration errors during parameter estimation proved especially useful.
Prior to the optimization process itself, the same acceleration measurements could be used to fit an initial guess of the lift and drag curves.
Such a plot can be seen in Fig.~\ref{fig:static_curves}.
\subsection{Model validation}
\begin{figure}
	\centering
	\includegraphics[width=0.47\linewidth]{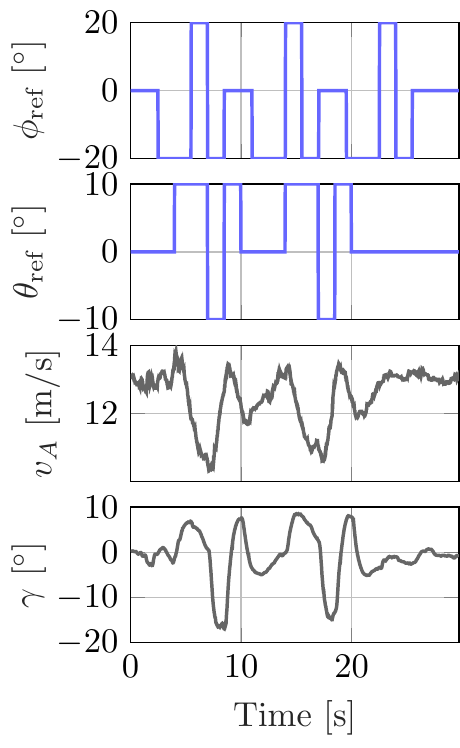}
	\hfill\includegraphics[width=0.49\linewidth]{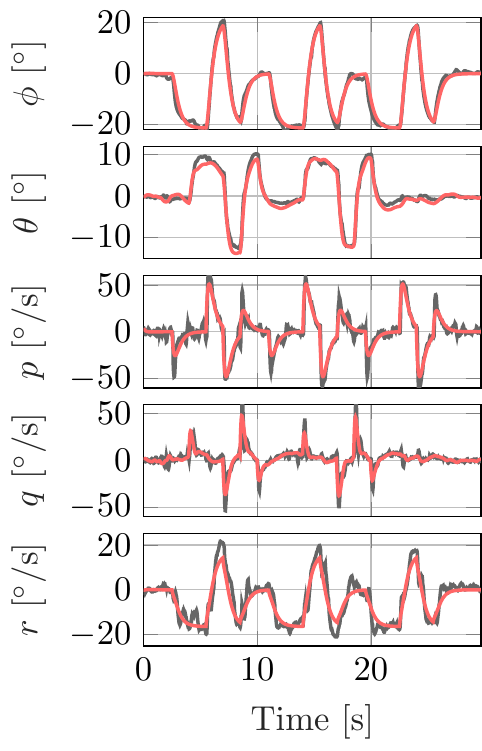}
	\caption{Inputs (left) and outputs (right) for a coupled validation experiment on the closed-loop attitude dynamics. (blue is the input signal, red is the simulated output)}
	\label{fig:idValCoupled-2}
\end{figure}

\begin{figure}
	\centering
	\includegraphics[width=0.45\linewidth]{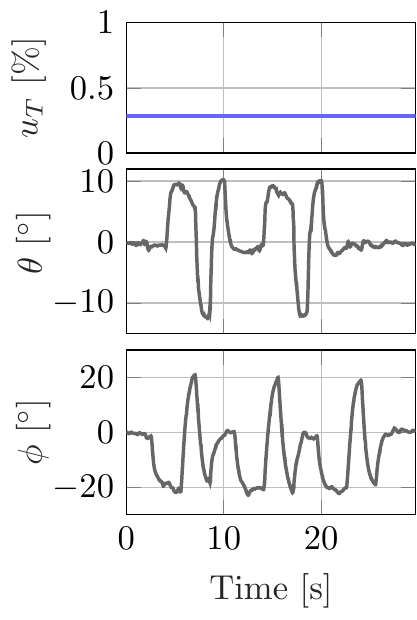}
	\hfill\includegraphics[width=0.49\linewidth]{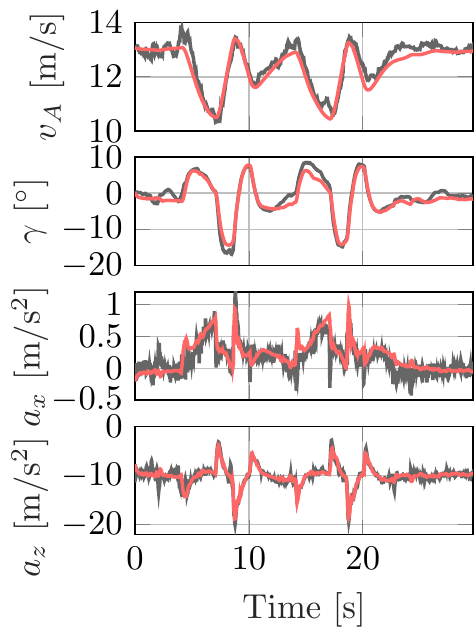}\hfill
	\caption{Inputs (left) and outputs (right) for a coupled validation experiment on the open-loop dynamics. (blue is the input signal, red is the simulated output)}
	\label{fig:idValCoupled-1}
\end{figure}

\begin{figure}
	\centering
	\includegraphics[width=0.36\linewidth]{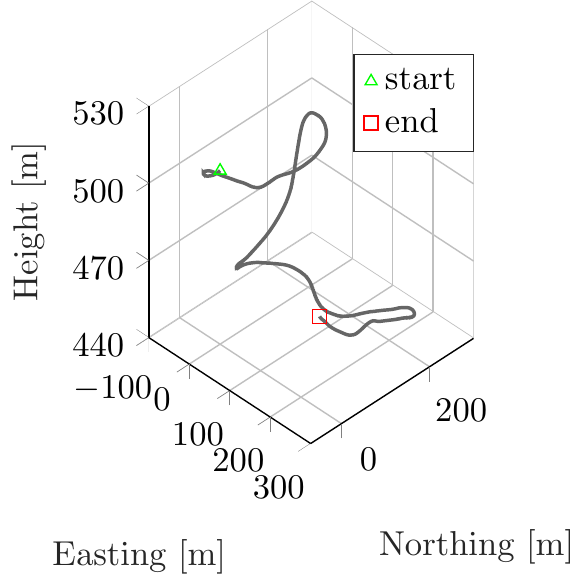}
	\includegraphics[width=0.62\linewidth]{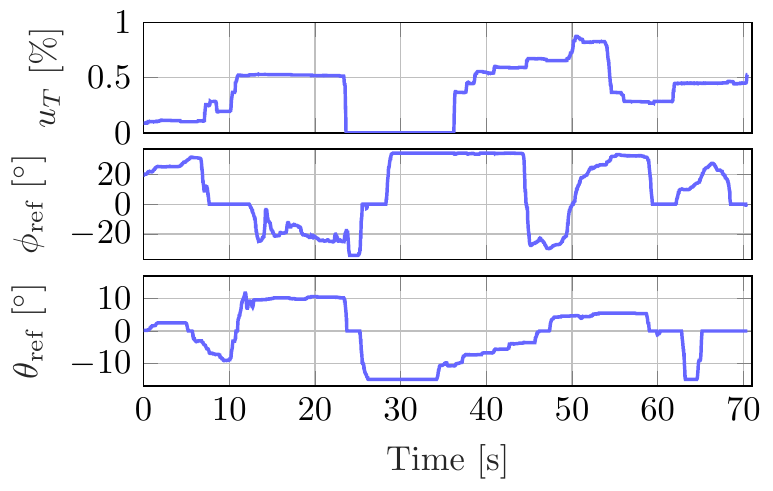}
	\caption{Flight path of the Techpod UAV (left) and corresponding full-model inputs (right) during a \emph{free-form} validation experiment.}
	\label{fig:idValFree-in}
\end{figure}

\begin{figure}
	\centering
	\includegraphics[width=0.99\linewidth,trim={0.1cm 0 0 0},clip]{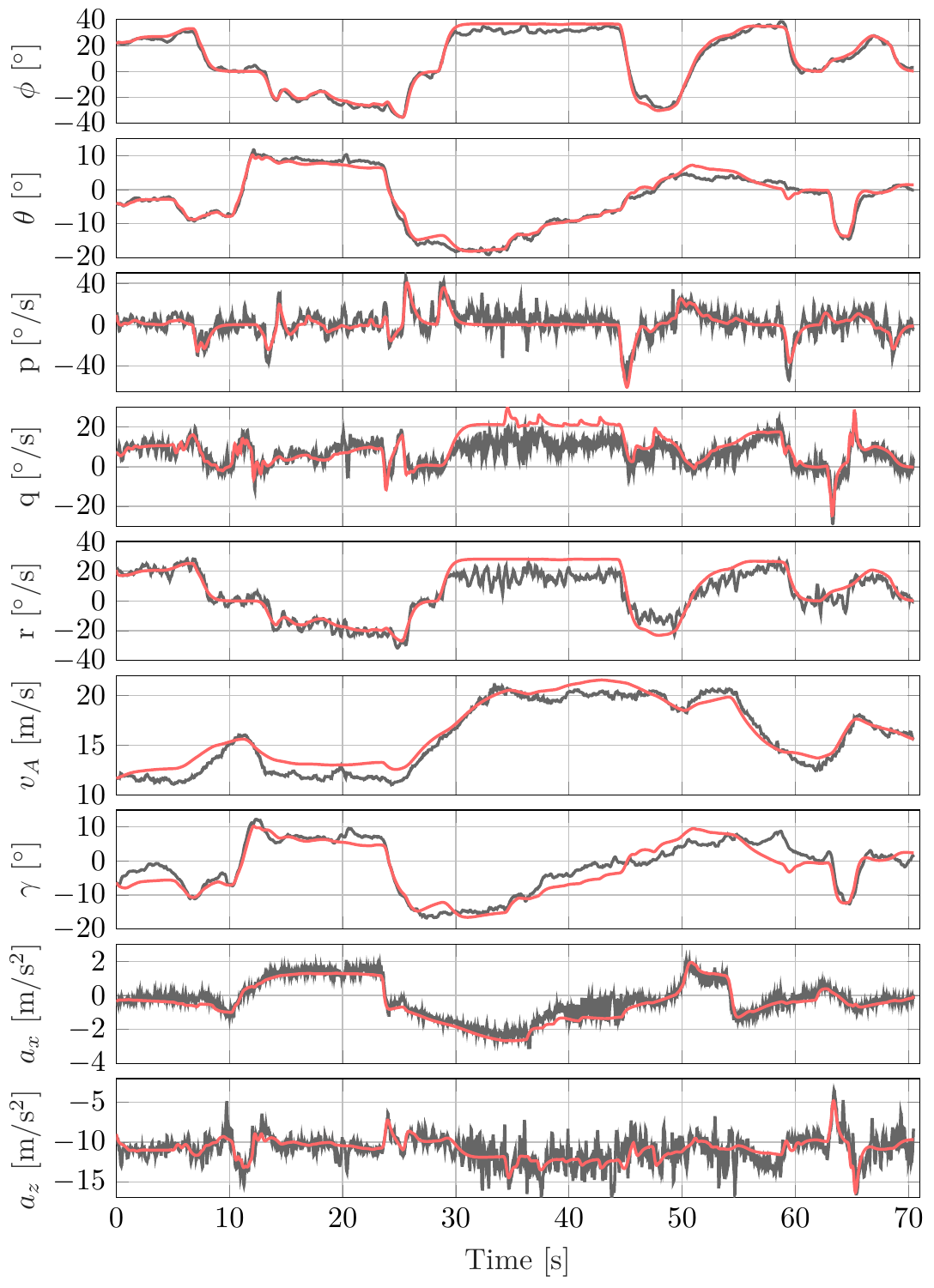}\hfill
	\caption{Free-form open-loop simulation of the combined, identified model (red) compared to flight data.}
	\label{fig:idValFree-out1and2}
\end{figure}

After optimizing the model parameter estimates of the two model structures, the models were validated on data not used within the training.
Figures~\ref{fig:idValCoupled-2} and~\ref{fig:idValCoupled-1} show a representative validation of a coupled pitching and rolling maneuver for the closed- and open-loop dynamics, respectively.
%
%
%
All outputs are well matched to the flight data.
Table~\ref{tab:RMSE_val} displays the average Root Mean Squared Error (RMSE) for each output signal over all validation sets.
\begin{table}[h]
\centering
	\caption{Average Root Mean Squared Error (RMSE) over all validation sets.}
	\begin{tabular}{!{\vrule width 0.5pt}ccc!{\vrule width 0.5pt}ccc!{\vrule width 0.5pt}}
	\multicolumn{3}{c}{$\mathbf{y}_\text{CL}$} & \multicolumn{3}{c}{$\mathbf{y}_\text{OL}$}\\\noalign{\hrule height 0.5pt}
	\rowcolor[gray]{0.9}Signal & RMSE & Unit & Signal & RMSE & Unit \\\noalign{\hrule height 0.5pt}
	$\phi$ & 1.610 & \SI{}{\degree} & $v_A$ & 0.424 & \SI{}{\meter\per\second}\\
	\rowcolor[gray]{0.9}$\theta$ & 0.921 & \SI{}{\degree} & $\gamma$ & 1.680  & \SI{}{\degree}\\
	$p$ & 5.140 & \SI{}{\degree\per\second} & $a_x$ & 0.217 & \SI{}{\meter\per\square\second}\\
	\rowcolor[gray]{0.9}$q$ & 3.390 & \SI{}{\degree\per\second} & $a_z$ & 0.660 & \SI{}{\meter\per\square\second}\\
	$r$ & 2.650 & \SI{}{\degree\per\second} & & & \\\noalign{\hrule height 0.5pt}
	\end{tabular}
	\label{tab:RMSE_val}
\end{table}
As both identified models are to be propagated simultaneously within the horizon of the MPC, a subset of \emph{free-form} flight data was used to \emph{test} the fully integrated model in open-loop simulation.
Figures~\ref{fig:idValFree-in} and~\ref{fig:idValFree-out1and2} show a comparison of one such simulation against over \SI{1}{\minute} of flight data.
Despite not being trained or validated with the combined model, the results show good tracking - validating the decoupled (open-loop vs. closed-loop) modeling assumptions made within the identification.
Notably, the largest errors within the experiment were seen during extended maximum roll angle commands while simultaneously flying at airspeeds exceeding the identified state-space -- suggesting, in particular, that the model structure for $q$ and $r$ dynamics may begin to break down near the boundary of the identified flight envelope.
For more aggressive flight with higher roll angles or airspeeds, these unmodeled effects would need further consideration.
\section{CONTROL FORMULATION}
In this section we formulate a high-level Nonlinear Model Predictive Controller (NMPC) for multi-objective guidance of the UAV, embedding the path following problem for optimization within the horizon, stabilizing airspeed, and considering soft constraints on the angle of attack.
\subsection{Path following}
Dubins Aircraft segments (lines and arcs in 3D)~\cite{beard2013implementing} can be used to describe the majority of desired flight maneuvers in a typical fixed-wing UAV mission.
Further, using simple paths such as arcs and lines allows spatially defined path following, independent of time (or consequently speed), a useful quality when only proximity to the track is desired and winds can significantly change the ground velocity.
For the remainder of the section, we will consider Dubins segments as path inputs to the high-level controller, though it should be noted that the path objective formulation is not limited to these; e.g. within this work we have also incorporated a special case of the Dubins arc, the common unlimited loiter circle.
\subsubsection{Path geometry}\label{sec:path_geom}
\begin{figure}
	\centering
	\includegraphics[width=0.45\linewidth]{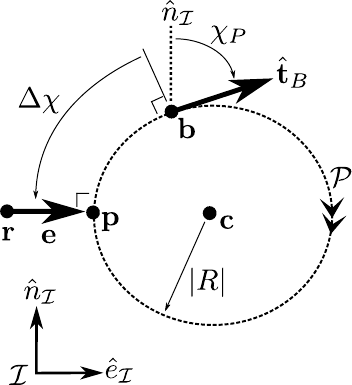}
	\hfill\includegraphics[width=0.5\linewidth]{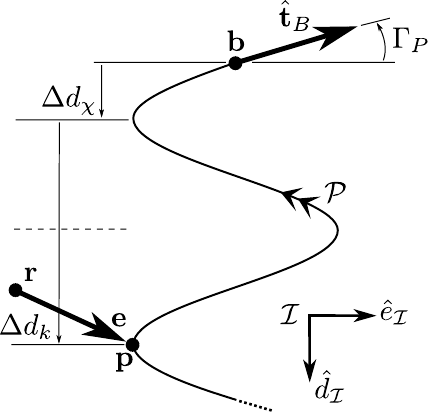}
	\caption{Lateral-directional (left) and longitudinal (right) \emph{arc} path geometry.}
	\label{fig:path-geom}
\end{figure}

A minimum set of path parameters are required to define each time independent segment type as follows:
\begin{itemize}
\item Dubins line: $\mathcal{P}\in\text{line}=\left[\mathbf{b},\chi_P,\Gamma_P\right]$
\item Dubins arc: $\mathcal{P}\in\text{arc}=\left[\mathbf{c},\pm R,\chi_P,\Gamma_P\right]$
\item Loiter unlim.: $\mathcal{P}\in\text{loit}=\left[\mathbf{c},\pm R\right]$
\end{itemize}
\noindent where $\mathbf{b}$ is the terminal point on a Dubins line, $\mathbf{c}$ is the center point of a Dubins arc (or loiter circle) at the terminal altitude, $\pm R$ is the arc (or loiter) radius with sign indicating clockwise (positive) or counter-clockwise (negative) direction, $\chi_P$ is the exit course for a Dubins arc or line, and $\Gamma_P$ is the inertial-frame elevation angle of a Dubins arc or line.
Figure~\ref{fig:path-geom} describes the geometry.
To date, fixed-wing guidance logic for tracking helix-type paths has largely been limited to `pose-in-time'-based definitions (e.g.~\cite{beard2013implementing,garcia-nmpcfixedwing_2011}), i.e. desired positions and orientations are prescribed in time from initially starting to follow a given path segment.
As the aircraft ground speed may change significantly over time, e.g. due to wind, we desire a time-independent formulation, defined only by spatial proximity to the path.
A unique spatially defined solution for the closest point on a \emph{line} in three-dimensions can be analytically calculated.
However, to avoid multiple solutions or the necessity of numerical methods when finding the closest point on the 3D \emph{arc} paths, we define an approximate of the closest point by decoupling the problem into lateral-directional and longitudinal planes.
We first consider the closest point in the lateral-directional plane to a circle with radius $|R|$ (with unique spatially defined solution, except on the center point), and subsequently choosing the nearest arc `leg' (assuming an infinite helix in the direction opposite the sign of the elevation angle).
In eq.~\eqref{eq:closest_point}, the $d$ component of the closest point on the path $p_d$ is calculated via summation of the terminal altitude $b_d$, the altitude deviation $\Delta d_\chi$ due to the angular distance $\Delta \chi$ from the exit point, and the altitude deviation $\Delta d_k$ corresponding rounded $k$ number of arc `legs' away from the terminal point; see Fig.~\ref{fig:path-geom}.
\begin{equation}\label{eq:closest_point}
\begin{array}{l}
\Delta d_\chi = \Delta \chi |R| \tan\Gamma_P \\
\Delta d_k = \operatorname{round}\left(\frac{d-\left(b_d+\Delta d_\chi\right)}{2\pi |R| \tan\Gamma_P}\right)2\pi |R| \tan\Gamma_P\\
p_d = b_d + \Delta d_\chi + \Delta d_k \\ 
\end{array}
\end{equation}
\subsubsection{Lateral-directional guidance}\label{sec:lat_guide}
\begin{figure}
	\centering
	\includegraphics[width=0.55\linewidth]{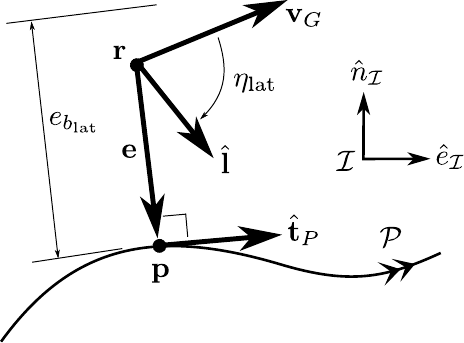}
	\caption{Lateral-directional guidance logic.}
	\label{fig:lat-guidance}
\end{figure}

As fixed-wing aircraft behave dissimilarly in lateral-directional and longitudinal states, we also decouple the guidance objectives.
The lateral track-error is defined:
\begin{equation}\label{eq:e_lat}
\begin{array}{l}
e_\text{lat} =  \bar{t}_{P_n}\left(p_e - r_e\right) - \bar{t}_{P_e} \left(p_n - r_n\right) \\
\end{array}
\end{equation}
\noindent where $\bar{t}_{P_n}=\frac{t_{P_n}}{\|\left[t_{P_n},t_{P_e}\right]\|}$ and $\bar{t}_{P_e}=\frac{t_{P_e}}{\|\left[t_{P_n},t_{P_e}\right]\|}$, for $\|\left[t_{P_n},t_{P_e}\right]\|\neq 0$.
The unit path tangent $\hat{\mathbf{t}}_P$ is defined from the current path parameters $\mathcal{P}_\text{cur}$, specifically $\chi_P$ and $\Gamma_P$.
One approach to the path following problem is to minimize the track-error itself along with the error between the aircraft course angle, intertial flight path angle, and the path tangent, e.g. $\chi-\chi_P$, as designed in the 2D case presented in our previous work~\cite{Stastny_GNC2017}, as well as other works~\cite{kang2006design,kang2009linear,yang2009adaptive}.
However, the NMPC's internal model also includes airspeed in an open-loop formulation, requiring simultaneous reference tracking.
This dichotomy was found to present a challenge in properly defining a time-independent and velocity independent path following objective, as large track-errors would induce increased airspeed commands in an attempt to quickly reduce the larger cost in the shortest time within the horizon.
Rather than attempt to define a complicated prioritized approach to weighting these two competing objectives, we instead embed unified, speed independent guidance logic, incorporating both directional and position errors into one lateral-directional and one longitudinal error term.
Augmentation of the NMPC internal model to include the guidance formulation has also previously been explored~\cite{garcia-nmpcfixedwing_2011,Stastny_GNC2015,Stastny_JDSMC2014}; however, in these approaches, the analytic guidance law was used to generate attitude references within the control horizon, commands we wish our high-level NMPC to allocate itself.
We therefore propose in this work to leave the control allocation open-ended for the nonlinear optimization to solve in real-time, while providing the NMPC with an error angle in the objective which, when minimized, results in convergence to the path.
The lateral-directional guidance error is formulated as the error angle $\eta_\text{lat}$ from a \emph{look-ahead} (or line-of-sight) guidance approach, commonly used in high-level lateral-directional position control for fixed-wing UAVs, see eq.~\eqref{eq:look_ahead} and Fig.~\ref{fig:lat-guidance}.
Specifically we formulate our look-ahead vector $\hat{\mathbf{l}}$ in a similar manner to the formulation found in~\cite{Cho_2015_NPFG}, though it should be noted that several similar formulations exist, e.g. \cite{park2007performance,L2plus}.
\begin{equation}\label{eq:look_ahead}
\begin{array}{l}
\hat{\mathbf{l}}=\left(\begin{matrix}
l_n \\
l_e\end{matrix}\right)=\left(\begin{matrix}
\left(1-\theta_{l_\text{lat}}\right)\bar{t}_{P_n}+\theta_{l_\text{lat}}\bar{e}_n \\
\left(1-\theta_{l_\text{lat}}\right)\bar{t}_{P_e}+\theta_{l_\text{lat}}\bar{e}_e
\end{matrix}\right)
\end{array}
\end{equation}
\noindent where $\bar{e}_n=\frac{{e}_n}{\|\left[{e}_n,{e}_e\right]\|}$ and $\bar{e}_e=\frac{{e}_e}{\|\left[{e}_n,{e}_e\right]\|}$, for $\|e_n,e_e\|\neq0$, and $\mathbf{e}=\mathbf{p}-\mathbf{r}$.
$\theta_{l_\text{lat}}$ is a mapping function for the lateral-directional track-error, equal to $1$ at the track-error boundary $e_{b_\text{lat}}$ and $0$ when $e_\text{lat}=0$.
We choose a quadratic shape for $\theta_{l_\text{lat}}$ such that beyond the track-error boundary a perpendicular approach to the path is demanded, and at the track-error boundary, the commanded direction begins to transition smoothly towards the unit tangent vector on the path:
\begin{equation}
\theta_{l_\text{lat}}=-e^\prime_\text{lat}\left(e^\prime_\text{lat}-2\right)
\end{equation}
\noindent where the normalized and saturated lateral-directional track-error is defined:
\begin{equation}
e^\prime_\text{lat}=\operatorname{sat}\left(|e_\text{lat}|/e_{b_\text{lat}},0,1\right)
\end{equation}
\noindent and $\operatorname{sat}\left(\cdot,\text{min},\text{max}\right)$ is a saturation function.
Borrowing a similar effect from the developments in~\cite{L2plus}, the track-error boundary is defined in an adaptive way with respect to the current ground speed $e_{b_\text{lat}}=\|\mathbf{v}_{G_\text{lat}}\|T_{b_\text{lat}}$, for $\|\mathbf{v}_{G_\text{lat}}\|\neq 0$, where $T_{b_\text{lat}}$ is a tuning constant, varying the steepness of the look-ahead vector mapping on approach to the path, and $\mathbf{v}_{G_\text{lat}}=\left[v_{G_n},v_{G_e}\right]^T$.
In this work, we limit the minimum track error bound $e_{b_\text{lat}}$ with a smooth piecewise relationship to a minimum ground speed, arbitrarily set to \SI{1}{\meter\per\second}, see eq.~\eqref{eq:e_b}.
\begin{equation}\label{eq:e_b}
e_{b_\text{lat}}=\begin{cases}
\|\mathbf{v}_{G_\text{lat}}\|T_{b_\text{lat}} & \|\mathbf{v}_{G_\text{lat}}\|>1 \\
\frac{1}{2}T_{b_\text{lat}}\left(1+\|\mathbf{v}_{G_\text{lat}}\|^2\right) & \text{else}
\end{cases}
\end{equation}
The final guidance objective is defined as the error angle between the aircraft ground speed vector and the look-ahead vector.
\begin{equation}
\eta_\text{lat} = \operatorname{atan2}\left(l_e,l_n\right) - \operatorname{atan2}\left(\dot{e},\dot{n}\right)
\end{equation}
\noindent where $\operatorname{atan2}$ is the four quadrant arctangent operator, and $\eta_\text{lat}$ should be wrapped to remain within $\pm \pi$.
Note this discrete switch at \SI{180}{\degree} entails an instability point within the guidance formulation when propagated within the horizon; e.g., in the case that the aircraft is traveling on the track in the opposite of the desired direction.
Careful initialization of the NMPC horizon should be considered to ensure operation outside of some range of this condition.
A hard end term constraint bounding the aircraft to find trajectories leading away from this zone would also be advised, however, at this point, the ACADO framework~\cite{Houska2011a} (used for auto-generation of optimized C code for real time control in this paper) does not support externally defined constraints.
In lieu of this, some handling of this case outside of the NMPC (e.g. shifting the track away when close to this configuration) is possible.
One further consideration is the limitation of maintaining an airspeed greater than the wind speed.
While the present guidance formulation will not fall into a singularity, the resulting guidance commands may be erroneous (e.g. when the $\operatorname{atan2}$ function on ground speed is very near zero on both input arguments), and this work does not further consider their influence on the NMPC's corresponding objective cost, apart from the insight that some objective weight retuning was found to required.
However, one may look to appropriate guidance enhancements for these particular conditions in, e.g., the formulation presented in~\cite{Furieri_ACC2017}.
\subsubsection{Longitudinal guidance}
We approach longitudinal guidance in a slightly different manner than that of the lateral-directional, as longitudinal fixed-wing states do not have the full range of their counterparts in the 2D, horizontal plane, and are non-symmetric in climbing and sinking flight performance.
We define the desired (on-track) vertical velocity $\dot{d}_P=\|\mathbf{v}_G\|t_{P_d}\in\left(\dot{d}_\text{clmb},\dot{d}_\text{sink}\right)$ corresponding to the path elevation and current ground speed, further bounded by the maximum climb rate $\dot{d}_\text{clmb}$ and sink rate $\dot{d}_\text{sink}$.
Depending on the sign and magnitude of the longitudinal track-error $e_\text{lon}=p_d-r_d$, a vertical velocity setpoint $\dot{d}_\text{sp}$ is modulated in an asymmetric manner between the bounds of maximum sinking and climbing using a quadratic look-ahead mapping function similarly defined to that within the lateral guidance, Sec.~\ref{sec:lat_guide}.
\begin{equation}\label{ddot_sp}
\dot{d}_\text{sp} = \Delta\dot{d}\theta_{l_\text{lon}} + \dot{d}_P
\end{equation}
\noindent with look-ahead mapping $\theta_{l_\text{lon}} = -e_\text{lon}^\prime\left(e_\text{lon}^\prime-2\right)$ and normalized track-error $e_\text{lon}^\prime = \operatorname{sat}\left(|e_\text{lon}/e_{b_\text{lon}}|,0,1\right)$, and track-error boundary defined similar to the lateral-directional:
\begin{equation}\label{eq:e_b_lon}
e_{b_\text{lon}}=\begin{cases}
T_{b_\text{lon}} |\Delta\dot{d}| & |\Delta\dot{d}|>1 \\
\frac{1}{2}T_{b_\text{lon}}\left(1+{\Delta\dot{d}}^2\right) & \text{else}
\end{cases}
\end{equation}
\noindent with $\Delta \dot{d}$ defined for climbing or sinking:
\begin{equation}\label{eq:delta_ddot}
\Delta \dot{d}=\begin{cases}
\Delta \dot{d}_\text{clmb} & e_\text{lon}<0 \\
\Delta \dot{d}_\text{sink} & \text{else}
\end{cases}
\end{equation}
\noindent where $\Delta \dot{d}_\text{clmb} = -\dot{d}_\text{clmb}-\dot{d}_P$ and $\Delta \dot{d}_\text{sink} = \dot{d}_\text{sink}-\dot{d}_P$.
The resultant guidance error term is formulated as the vertical velocity offset, normalized by the range of climbing and sinking rates.
\begin{equation}\label{eq:eta_lon}
\begin{array}{l}
\eta_\text{lon} = \frac{\dot{d}_\text{sp} - \dot{d}}{\dot{d}_\text{clmb}+\dot{d}_\text{sink}}
\end{array}
\end{equation}
\noindent where $\eta_\text{lon}$, though not a true angular error as in the lateral-directional case, is then mostly defined between $[-1,1]$ (except with large vertical velocity deviations). 
Another purpose of the present formulation is to allow for high horizontal wind scenarios where we may still be able to climb or sink to a desired altitude, despite flying at close to zero horizontal ground speed, something an angle based guidance objective would not readily handle.
\subsubsection{Switching conditions}
\label{sec:sw}
Terminal conditions for Dubins arc segments require the aircraft to be within some acceptance radius $R_\text{acpt}$ of the segment terminal point $\mathbf{b}$ (proximity), traveling within an acceptance angle $\eta_\text{acpt}$ of the exit course $\chi_P$ (bearing), and beyond the terminal point $\mathbf{b}$ in the path axis (travel).
Only the travel condition is set for line segments to avoid runaway behavior when the other conditions are missed due to e.g. the path being commanded while the aircraft is not already close to the track and correct orientation.
No terminal condition is set for unlimited loiter circles.
Switching conditions are summarized in eq.~\eqref{eq:sw_condition}, and shown graphically in Fig.~\ref{fig:path-geom}.
\begin{equation}\label{eq:sw_condition}
\begin{array}{ll}
\|\mathbf{r}-\mathbf{b}\|<R_\text{acpt} & \text{(proximity)} \\[0.05cm]
\mathbf{v}_G\cdot\hat{\mathbf{t}}_B>\cos\eta_\text{acpt} & \text{(bearing)} \\[0.05cm]
\left(\mathbf{r}-\mathbf{b}\right)\cdot\hat{\mathbf{t}}_B>0 & \text{(travel)}
\end{array}
\end{equation}
\noindent where $\hat{\mathbf{t}}_B$ is the unit tangent at the terminal point $\mathbf{b}$ of the current path $\mathcal{P}_{cur}$. Note for Dubins arcs, $\mathbf{b}$ must be calculated from the arc center $\mathbf{c}$ and exit course $\chi_P$.
%

As in~\cite{Kamel2017_ROSBOOK,Stastny_GNC2017}, a switching state $x_{sw}$ is defined and augmented to the model, with dynamic shown in eq.~\eqref{eq:x_sw}.
Then, $x_{sw}$ defines the desired path in the queue to follow internally within the horizon.
\begin{equation}
\dot{x}_{sw}=\begin{cases} 
      \rho & \text{terminal conditions met}\quad \| \quad x_{sw}>\text{threshold} \\
      0 & \text{else}
   \end{cases}
   \label{eq:x_sw}
\end{equation}
\noindent where $\rho$ is an arbitrary constant.
\subsection{Optimal control problem}
We use the ACADO Toolkit~\cite{Houska2011a} for the generation of a fast C code based nonlinear solver and implicit Runge-Kutta integration scheme.
A direct multiple shooting technique is used to solve the optimal control problem (OCP), where dynamics, control action, and inequality
constraints are discretized over a time grid of a given horizon length $N$.
A boundary value problem is solved within each interval and additional continuity constraints are imposed.
Sequential Quadratic Programming (SQP) is used to solve the individual QPs, using the active set method implemented in the \emph{qpOASES}\footnote{\url{http://www.qpOASES.org/}} solver.
The OCP takes the continuous time form:
\begin{equation} \label{eq:nmpc_opt}
\begin{aligned}
\min_{\mathbf{x},\mathbf{u}} &\
\int_{t=0}^{T} \left(\left(\mathbf{y}(t) - \mathbf{y}_\text{ref}(t)\right)^{T}\mathbf{Q}_y\left(\mathbf{y}(t) - \mathbf{y}_\text{ref}(t)\right)\right.\\
& \left.+ \left(\mathbf{z}(t) - \mathbf{z}_\text{ref}(t)\right)^{T}\mathbf{R}_z\left(\mathbf{z}(t) - \mathbf{z}_\text{ref}(t)\right)\right) dt\\
& + \left(\mathbf{y}(T) - \mathbf{y}_\text{ref}(T)\right)^{T}\mathbf{P}\left(\mathbf{y}(T) - \mathbf{y}_\text{ref}(T)\right)\\[0.2cm]
&\begin{array}{lll}
\text{subject to} & \dot{\mathbf{x}} = f(\mathbf{x},\mathbf{u}), \\
& \mathbf{u}(t) \in \mathbb{U}, \\
& \mathbf{x}(0) = \mathbf{x}\left( {t_0}\right)
\end{array}
\end{aligned}
\end{equation}
\noindent where control vector $\mathbf{u}=\left[u_T,\phi_\text{ref},\theta_\text{ref}\right]^T$ and state vector $\mathbf{x}=\left[\mathbf{r}^T,\mathbf{v}_V^T,\mathbf{\Theta}^T,\bm{\omega}^T,\delta_T,x_{sw}\right]^T$. $\mathbf{Q}_y$, $\mathbf{R}_z$, and $\mathbf{P}$ are state, control, and end-term non-negative diagonal weighting matrices.
State and control-dependent output vectors are compiled from path following objectives, airspeed stabilization, rate damping, and soft constraints:
\begin{equation}
\begin{array}{l}
\mathbf{y}=\left[\bm{\eta}^T,v_A,\bm{\omega}^T,\alpha_\text{soft}\right]^T \\
\mathbf{z}=\left[\dot{\delta}_T,\mathbf{u}^T\right]^T
\end{array}
\end{equation}
\noindent where $\bm{\eta}=\left[\eta_\text{lat},\eta_\text{lon}\right]^T$, and $\alpha_\text{soft}$ is a soft constraint on the angle of attack (to be defined in Sec.~\ref{sec:soft_const}).
\subsubsection{Feed-forward terms}
Note that penalization of the attitude references and throttle input is necessary to avoid bang-bang control behavior.
The selection of trim values, however, is important to avoid lowered performance in other objectives.
Constant trim values (for level-cruise flight) are used as control output references $\mathbf{z}_\text{ref}=\left[0,u_{T_\text{trim}},0,\theta_\text{trim}\right]$.
However, it was found that a feed-forward calculation for an approximate roll angle reference $\phi_\text{ref}$ improves path following objective performance, as roll angle trims for turning flight have large offsets from level flight.
The feed-forward term is calculated and subtracted from the roll angle reference $\phi_\text{ref}$ throughout the horizon for the final output $z_{\phi_\text{ref}}=\phi_\text{ref}-\phi_\text{ff}$:
\begin{equation}
\phi_\text{ff}=\begin{cases}
\tan^{-1}\left(\frac{\|\mathbf{v}_{G_\text{lat}}\|^2}{g R}\right)\frac{1+\cos\left(\pi e^\prime_\text{lat}\right)}{2} & \mathcal{P}_\text{cur}\in\text{arc},\text{loit} \\
0 & \mathcal{P}_\text{cur}\in\text{line} \\
\end{cases}
\end{equation}
Note $\phi_\text{ff}$ is not a commanded value to be explicitly tracked, but only gives guidance when near the track (via the multiplied smooth trig function as a function of the normalized lateral track error $e^\prime_\text{lat}$), keeping the weighted roll trim closer to the region of the optimal solution.
\subsubsection{Soft constraints}\label{sec:soft_const}
As the NMPC is also required to stabilize the open-loop dynamics of the vehicle, inappropriate commands could lead to a stall of the aircraft.
To mitigate the potential for stall, we include a soft constraint on the angle of attack $\alpha$, keeping zero cost within the ``safe" range, and quadratically increasing cost outside of these minimum $\alpha_-$ and maximum $\alpha_+$ bounds.
A transition zone is defined by $\Delta\alpha$ to allow tuning of the constraint's steepness.
\begin{equation}\label{eq:alpha_soft}
\alpha_\text{soft}=\begin{cases}
\left(\frac{\alpha-\left(\alpha_+-\Delta\alpha\right)}{\Delta\alpha}\right)^2 & \alpha>\alpha_+ \\
0 & \alpha_+\geq\alpha>\alpha_- \\
\left(\frac{\alpha-\left(\alpha_-+\Delta\alpha\right)}{\Delta\alpha}\right)^2 & \text{else}
\end{cases}
\end{equation}
We have chosen soft constraints over hard constraints for several reasons.
Namely, as the NMPC is operating on the high-level state-space, angle of attack rates and other fast modes are not modeled and likely not to be regulated well using only the attitude commands at its disposal (stall prevention is traditionally a low-level control problem).
We thus only consider future prevention through foresight into the horizon, and allow momentary violations of the soft bounds during abrupt events like an actuator failure, strong gust, or poor initialization of the controller, instances where a hard constraint could result in either no control solution or more iteration steps leaving the low-level controller without commands for some time.
Note that other works have incorporated similar soft constraints within the MPC framework at the same time combining them with hard constraints~\cite{Kamel2017_RobustCollisionAvoidMAVNMPC}, in the cited example for the purpose of collision avoidance.
A similar option could be explored for stall prevention in future work.
\section{FLIGHT EXPERIMENTS}
As the primary focus of this paper was the experimental implementation and validation of the proposed guidance methods, we present several indicative flight experiments, providing insights gained from the field experience, and omit simulation studies for brevity.
\subsection{Hardware setup}
\begin{figure}
	\centering
	\includegraphics[width=0.42\linewidth]{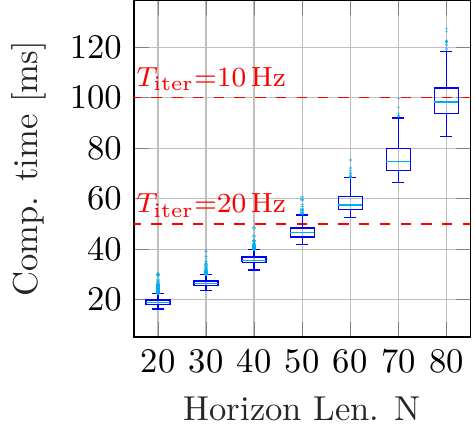}
	\hfill\includegraphics[width=0.55\linewidth]{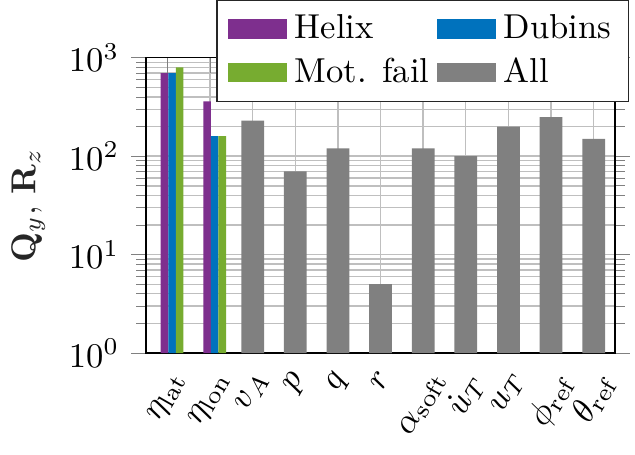}
	\caption{ROS node bench-tested on an Intel\textsuperscript{\textregistered} UP board (Quad Core, \SI{1.92}{\giga\hertz} CPU, \SI{4}{\giga\byte} RAM) in HIL configuration with horizon step size $T_\text{step}$=\SI{0.1}{\second}. State measurement updates were randomized and a 3D Dubins arc segment was input to maximize computational load. Approx. \SI{1}{\minute} of data collected for each configuration. (left) Objective weights for the normalized error outputs during each flight experiment. (right)}
	\label{fig:comptime_weights}
\end{figure}

Low-level attitude stabilization was run at \SI{50}{\hertz} on the Pixhawk, state estimates from the EKF were transferred via UART connection at \SI{40}{\hertz} over MAVLink/MAVROS to an on-board computer running Robotic Operating System (ROS), where a wrapper node iterated the NMPC solver at a specified fixed time interval $T_\text{iter}$.
As future applications of NMPC based guidance approaches may include objectives such as high speed obstacle avoidance, and noting that fixed-wing platforms require some time/space to maneuver, it is important to examine achievable real-time horizon lengths for experimentation.
Hardware-in-the-loop (HIL) bench tests were performed for this purpose, see box plot results in Fig.~\ref{fig:comptime_weights}.
Here, \emph{computation time} is defined as the sum of all operations conducted within the ROS node (solve time, array allocation, waypoint management, etc.) for one iteration.
\subsection{Experimental results}
Objective weighting was kept mostly constant throughout all experiments, save for some minor tuning adjustments, see Fig.~\ref{fig:comptime_weights} for a comparison.
Note that the output error signals $\mathbf{y}_\text{ref}-\mathbf{y}$ and $\mathbf{z}_\text{ref}-\mathbf{z}$ were normalized by the expected error ranges for nominal flight in an attempt to improve intuition on relative weighting between signals.
Guidance parameters were fixed throughout all experiments, see Table~\ref{tab:guidance_params}.
Further insight into the specific weighting is elaborated within each experiments subsection.
%
%
\begin{table}
\centering
	\caption{Guidance parameters used for flight experiments.}
	\begin{tabular}{!{\vrule width 0.5pt}ccc!{\vrule width 0.5pt}ccc!{\vrule width 0.5pt}}
	\noalign{\hrule height 0.5pt}\rowcolor[gray]{0.9} Param & Value & Unit & Param & Value & Unit \\\noalign{\hrule height 0.5pt}
	$\Delta\alpha$ & 2 & \SI{}{\degree} & $T_{b_\text{lat}}$ & 1 & \SI{}{\second}\\
	\rowcolor[gray]{0.9} $\alpha_-$ & -3  & \SI{}{\degree} & $T_{b_\text{lon}}$ & 1 & \SI{}{\second}\\
	$\alpha_+$ & 8 & \SI{}{\degree} & $\dot{d}_\text{climb}$ & 3.5 & \SI{}{\meter\per\second} \\
	\rowcolor[gray]{0.9} $R_\text{acpt}$ & 30 & \SI{}{\meter} & $\dot{d}_\text{sink}$ & 1.5 & \SI{}{\meter\per\second}\\
	$\eta_\text{acpt}$ & 15 & \SI{}{\degree} & & & \\\noalign{\hrule height 0.5pt}
	\end{tabular}
	\label{tab:guidance_params}
\end{table}
%

\subsubsection{Helix following}
\begin{figure}
	\centering
	\includegraphics[width=0.9\linewidth]{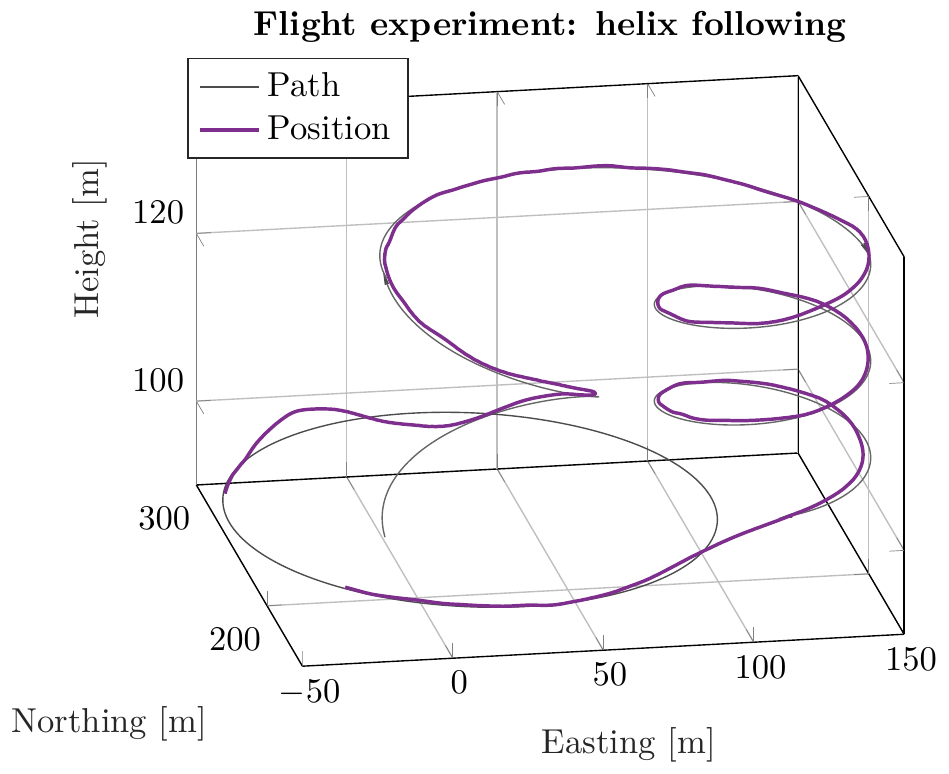}
	\caption{Techpod tracks a steep ascending (\SI{8}{\degree} incline) Dubins arc with a \SI{35}{\meter} radius, then summits on a large constant altitude arc before descending on another \SI{35}{\meter} radius arc (\SI{3}{\degree} glide).}
	\label{fig:exp-spiral-pos}
\end{figure}

\begin{figure}
	\centering
	\includegraphics[width=1\linewidth]{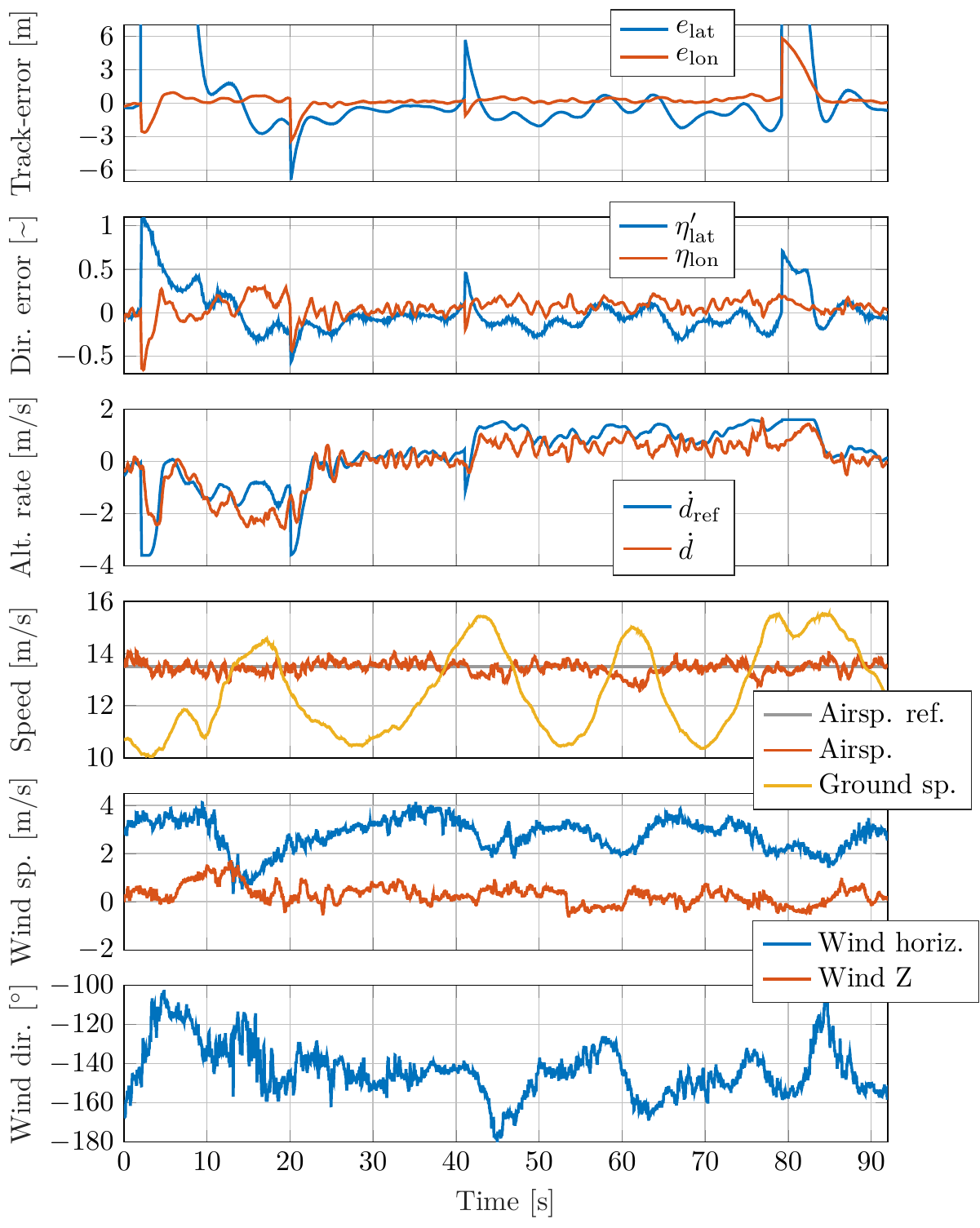}
	\caption{Guidance/track errors and air-, wind, and ground speeds during helix testing.}
	\label{fig:exp-spiral-err-speeds}
\end{figure}

To push the boundaries of the control horizon in a real world setting, a horizon length of $N=70$ was used with $T_\text{step}$=\SI{0.1}{\second}, corresponding to a \SI{7}{\second} horizon.
Control solutions were iterated at $T_\text{iter}$=\SI{0.1}{\second}, or \SI{10}{\hertz}.
Figures~\ref{fig:exp-spiral-pos} and~\ref{fig:exp-spiral-err-speeds} show path following and airspeed stabilization on steep ascending and descending Dubins arcs.
The arc radii were chosen to be just above the physical limit for the given roll angle constraints ($\pm$\SI{30}{\degree}) and commanded flight speed (\SI{13.5}{\meter\per\second}).
Final horizontal track-errors are kept within $\pm$\SI{2}{\meter} once settled to the path, and the vertical track-errors mostly regulated below $\pm$\SI{0.5}{\meter}.
More extensive flight testing on these tight and steep helix paths, especially when flying in wind, showed that appropriate relative tuning of the weights for the lateral-directional and longitudinal guidance errors is important.
Less relative weighting on the longitudinal path objective, at times, allowed significant altitude deviations on down-wind portions of the helix; though, once turning back into the wind the ground speed was lowered and the altitude error was again regulated.
The altitude deviations on the down-wind leg could become significant enough that the midpoint between helix legs was passed, causing a discrete switch to the lower leg within the horizon (recall the position-based helix logic in Sec.~\ref{sec:path_geom}).
While the latter point was later solved with a simple ``arc length traveled" logic to refuse \emph{previous} legs in each horizon, the prior required some investigation and subsequent adjustment to the objective weighting.
Convergence to this local minimum was in part due to the greater roll angle requirements for tracking the arc down-wind (faster ground speed), which reduced the vertical component of lift available for the steep climb and thus increased the required thrust and/or angle of attack (controlled with pitch), which would, in turn, cause greater deviations of these values from their constant objective references.
This, coupled with the higher weight on lateral-directional track error, induced lessened prioritization of the altitude error within the optimization.
Higher weighting on the longitudinal path objective (see~\ref{fig:comptime_weights}) resulted in the improved performance seen in Fig.~\ref{fig:exp-spiral-pos}.
%

\subsubsection{Connected Dubins segments}
\begin{figure}
	\centering
	\includegraphics[width=0.85\linewidth]{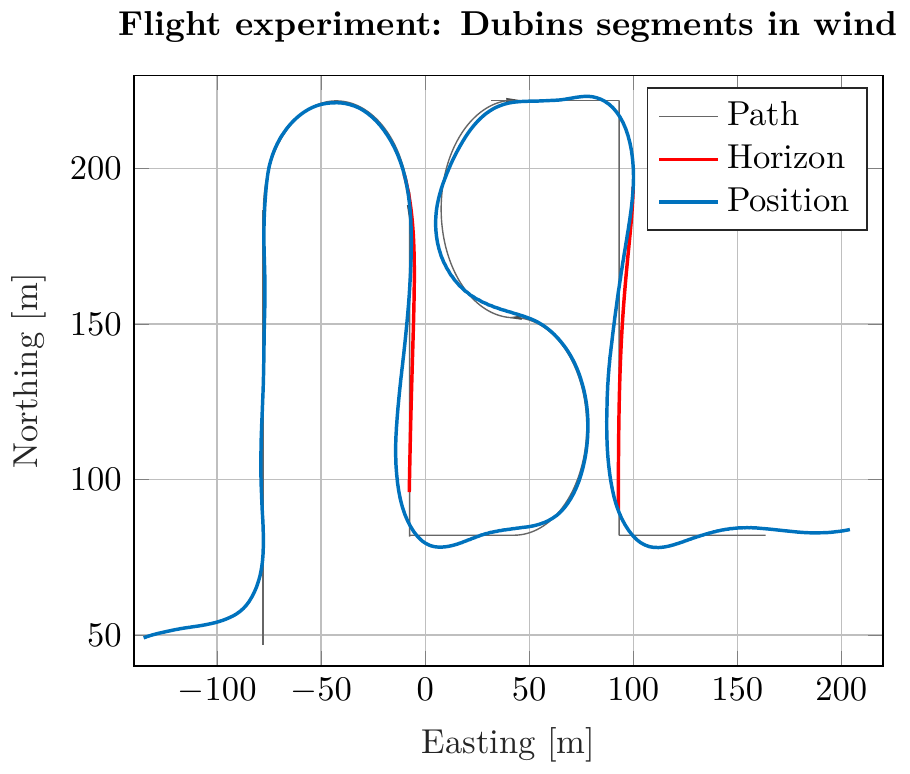}
	\caption{Techpod tracks connected Dubins arcs and lines in ca. \SI{5}{\meter\per\second} winds. The \emph{red} NMPC horizons showcase the planned trajectory converging to the straight segments before the horizon reaches the terminal conditions for switching to the next segment.}
	\label{fig:exp-asl-pos}
\end{figure}

\begin{figure}
	\centering
	\includegraphics[width=1\linewidth]{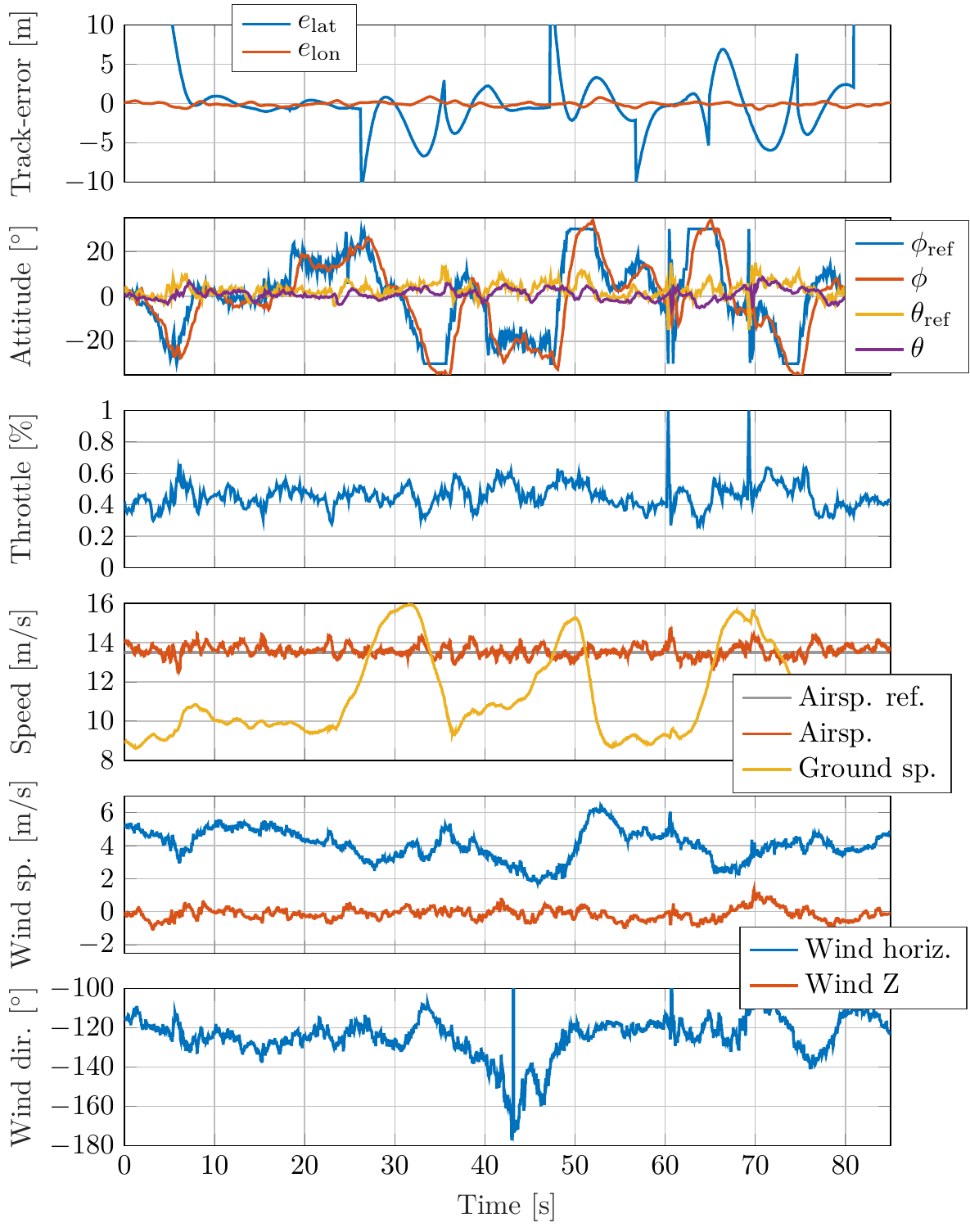}
	\caption{Track-errors and control inputs and air-, wind, and ground speeds during Dubins segment tracking in wind.}
	\label{fig:exp-asl-err-speeds}
\end{figure}

Figures~\ref{fig:exp-asl-pos} and~\ref{fig:exp-asl-err-speeds} show path following of connected Dubins lines and arcs.
As in the previous experiment, tight radii were commanded on each arc segment, and newly, the incorporation of line segments with \SI{90}{\degree} corners.
A \SI{7}{\second} horizon (length $N=70$, $T_\text{step}$=\SI{0.1}{\second}) with iteration rate \SI{10}{\hertz} was again used.
Note despite the $\sim$\SI{5}{\meter\per\second} wind and tight arcs, the horizontal track-errors are kept within $\pm$\SI{1}{\meter} once settled to a given set of smoothly connected path segments (see the left leg and top curvature of the \emph{A} in \emph{ASL}, Fig.~\ref{fig:exp-asl-pos}) and the vertical track-errors mostly regulated below $\pm$\SI{0.5}{\meter}.
Once the terminal conditions are met, the planned trajectory considers not only the current path, but also the next, allowing reduced tracking performance in the down-wind leg (also considering body-rate penalties) in order to reduce overshoot after the \SI{90}{\degree} turn onto the next.
One may notice the slightly noisy attitude and throttle reference signals.
Though in the present cases not having significant detrimental effect on the aircraft performance, it is worth noting the origin is primarily from the unfiltered (aside from a subtracted estimated bias) angular rate feedback to the NMPC.
Future consideration of some feedback or control output filtering may be warranted for particular signals, though care should be taken not to add undesired delays on the controller response.
%

\subsubsection{Motor failure}
\begin{figure}
	\centering
	\includegraphics[width=0.85\linewidth]{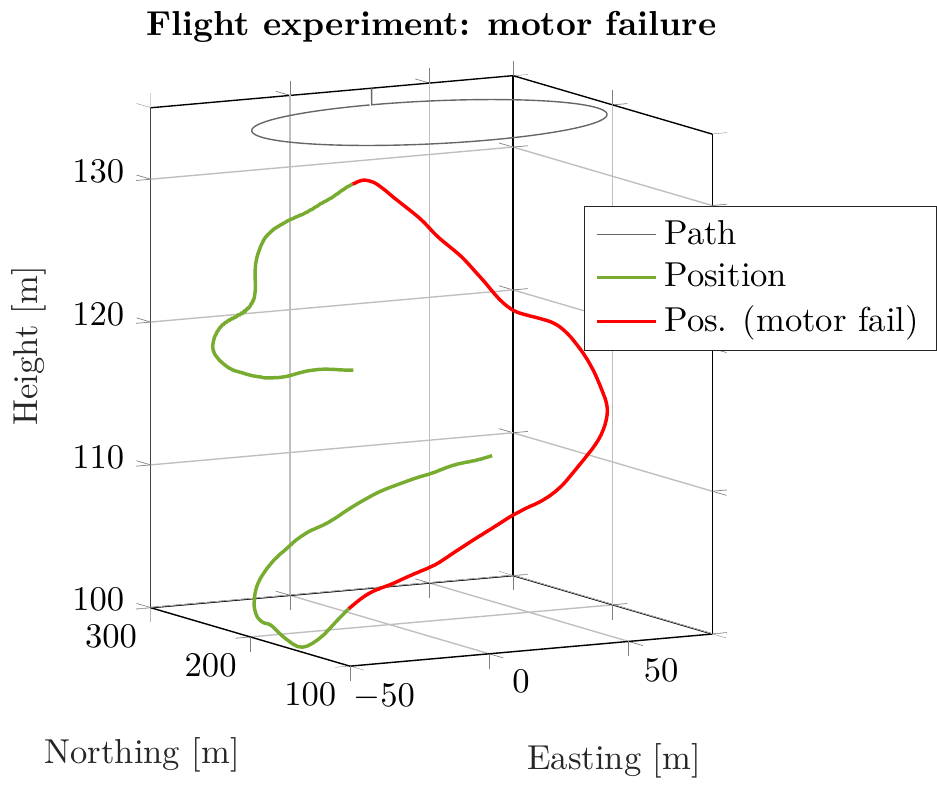}
	\caption{Techpod experiences a mock- motor failure during ascent to a loiter path.}
	\label{fig:exp-motfail-pos}
\end{figure}

\begin{figure}
	\centering
	\includegraphics[width=1\linewidth]{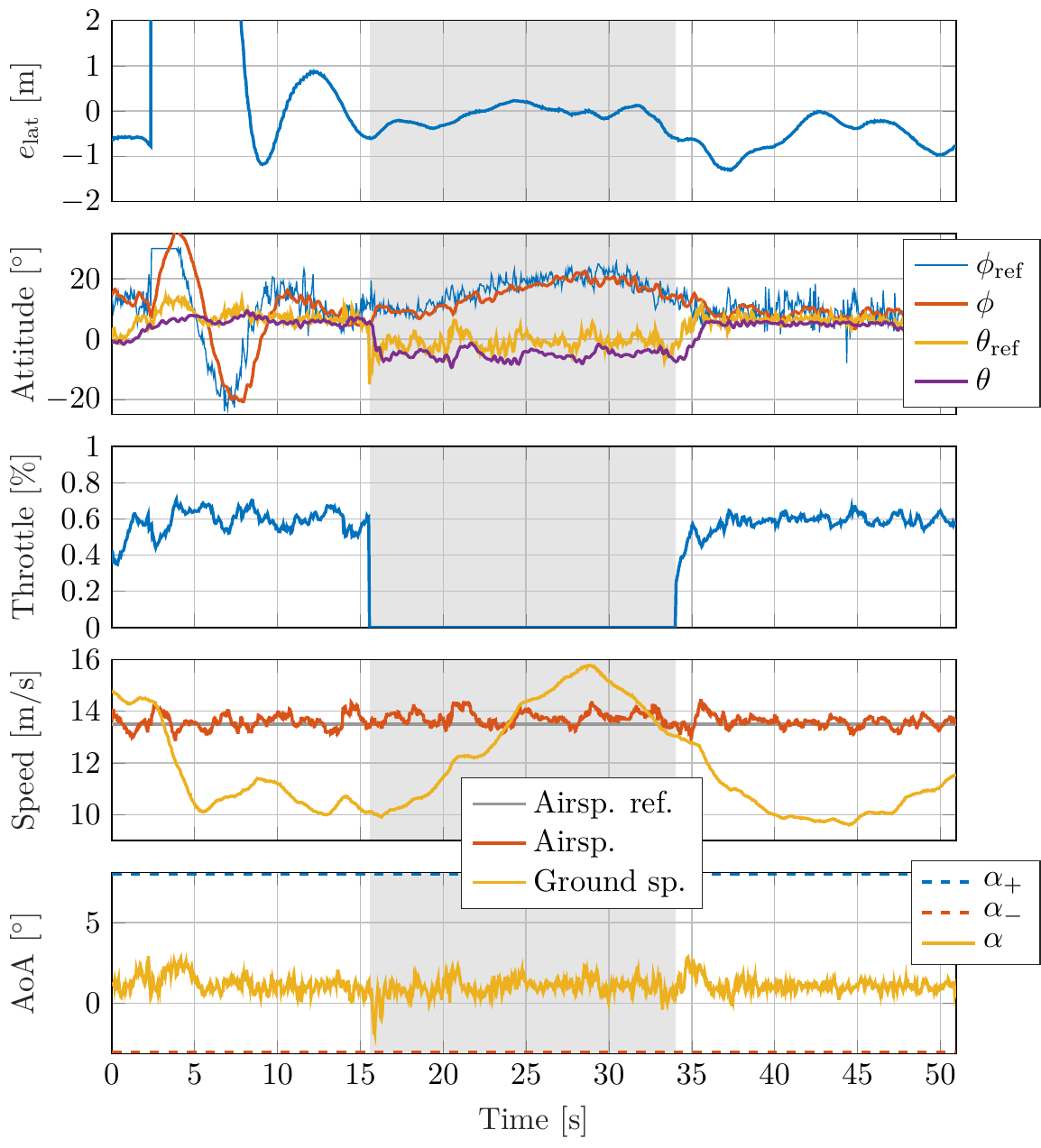}
	\caption{Lateral-directional track-error and control inputs (left) and airspeed, ground speed, and angle of attack (right) during a mock- motor failure.}
	\label{fig:exp-motfail-err-speeds}
\end{figure}

To explore potential fault tolerance of the designed guidance algorithm, we simulate a motor outage during a flight experiment.
This tests the NMPC's capability of reconfiguring the control allocation for the multi-objective problem in real-time.
We assume detection of a motor failure, generally (though perhaps not the exact type of failure), can be accomplished by monitoring the expected current draw (with respect to the throttle input) and comparing to a threshold value, a feature presently integrated in the custom flight software running on the Techpod UAV.
In this experiment, we change the horizon length to $N=40$, again with $T_\text{step}$=\SI{0.1}{\second}, a \SI{4}{\second} horizon.
However, with the reduced computational load (see Fig.~\ref{fig:comptime_weights}) we are able to increase the iteration rate to \SI{20}{\hertz} (or $T_\text{iter}$=\SI{0.05}{\second}).
A higher iteration rate allows faster feedback for the quicker dynamics expected to need mitigation in a motor failure situation.
Figures~\ref{fig:exp-motfail-pos} and~\ref{fig:exp-motfail-err-speeds} show the Techpod UAV ascending to a loiter circle, when the motor is cut at $t$=\SI{15.5}{\second}.
We simultaneously apply an arbitrarily large weight on the throttle input (e.g. $1e6$).
This causes the NMPC to reallocate the remaining control signals, in this case, immediately pitching down and stabilizing a glide at the commanded airspeed.
Notice that the lateral-directional track-error still remains below $\pm$\SI{1}{\meter}, as the optimization is able to maintain this particular tracking objective.
A brief, negative spike in the angle of attack $\alpha$ is seen at the time of the failure, but this is quickly returned to nominal values, and well within the bounds of the soft constraints (stopping just before the buffer zone).
At $t$=\SI{34}{\second}, the motor is reactivated, and the NMPC is similarly able to quickly reconfigure the control allocation and resume ascending to the loiter circle.
Despite the almost halving of the NMPC horizon, and the doubling of the iteration rate, no major retuning of the objective weights and parameters was necessary to maintain good performance.
\section{CONCLUSIONS \& FUTURE WORK}
In this work, we presented an approach for modeling and identification of the control augmented dynamics of a small fixed-wing UAV with a typical OTS autopilot in the loop, and further, utilized these dynamics within the internal model in the design of a high-level NMPC for simultaneous airspeed stabilization, 3D path following, and handling of soft angle of attack constraints.
The identified model structure demonstrated good predictive qualities, as shown with comparisons of open-loop simulation times on the order of tens of seconds, with respect to flight data. 
The designed high-level NMPC showed good performance for the multi-objective problem in a variety of experimental scenarios; in particular showcasing the benefit of explicit consideration of wind within the formulation and sufficiently long horizon times.
%

In future work, considering both airspeed and attitude within the low-level control loop would be advantageous, allowing a simplification of the control augmented model structure and, further, the possibility of increased horizon lengths and/or additional objectives such as obstacle avoidance and/or terrain constraints on autonomous landing.
\section*{Acknowledgment}
This research received funding from the Federal office armasuisse S+T under project number n$^\degree$050-45.
The authors would also like to thank Adyasha Dash for her contribution to the initial PX4 implementation of automated identification maneuvers extended and utilized within this work. 
\bibliographystyle{ieeetr}
\bibliography{lib}

\end{document}